\pdfoutput=1

\documentclass[11pt]{article}

\usepackage{EMNLP2023}

\usepackage{times}
\usepackage{latexsym}
\usepackage{makecell}
\usepackage{url}
\usepackage{CJKutf8}
\usepackage{subcaption}
\usepackage{graphicx}
\usepackage{subcaption}
\usepackage{todonotes}
\usepackage{amsmath}
\usepackage{booktabs}
\usepackage{multirow}
\usepackage{arydshln}

\usepackage{subcaption}
\usepackage{tcolorbox}

\usepackage[T1]{fontenc}

\usepackage[utf8]{inputenc}

\usepackage{microtype}

\usepackage{inconsolata}

\title{Towards Automatic Evaluation for Image Transcreation}

\author{Simran Khanuja$^{1}$\thanks{ denotes equal contribution} \quad Vivek Iyer$^{2}$\footnotemark[1] \quad Claire He$^{2}$ \quad \quad Graham Neubig$^{1}$ \\
  $^1$Carnegie Mellon University \quad $^{2}$University of Edinburgh \\
  \texttt{skhanuja@andrew.cmu.edu} \quad \texttt{vivek.iyer@ed.ac.uk} \\}

\begin{document}
\maketitle
\begin{abstract}
Beyond conventional paradigms of translating speech and text, recently, there has been interest in automated transcreation of \emph{images} to facilitate localization of visual content across different cultures. Attempts to define this as a formal Machine Learning (ML) problem have been impeded by the lack of automatic evaluation mechanisms, with previous work relying solely on human evaluation. In this paper, we seek to close this gap by proposing a suite of automatic evaluation metrics inspired by machine translation (MT) metrics, categorized into: \textbf{a)} \emph{Object-based}, \textbf{b)} \emph{Embedding-based}, and \textbf{c)} \emph{VLM-based}. Drawing on theories from translation studies and real-world transcreation practices, we identify three critical dimensions of image transcreation: \emph{cultural relevance}, \emph{semantic equivalence} and \emph{visual similarity}, and design our metrics to evaluate systems along these axes. Our results show that proprietary VLMs best identify \emph{cultural relevance} and \emph{semantic equivalence}, while vision-encoder representations are adept at measuring \textit{visual similarity}. Meta-evaluation across 7 countries shows our metrics agree strongly with human ratings, with average segment-level correlations ranging from \textbf{0.55-0.87}. Finally, through a discussion of the merits and demerits of each metric, we offer a robust framework for automated image transcreation evaluation, grounded in both theoretical foundations and practical application. Our code can be found here.\footnote{\url{https://github.com/simran-khanuja/automatic-eval-img-transcreation}}

\end{abstract}

\section{Introduction}
\label{sec:introduction}

\begin{figure}[!t]
    \centering
    \begin{subfigure}{0.45\columnwidth}
        \includegraphics[width=1.05\linewidth,height=\linewidth]{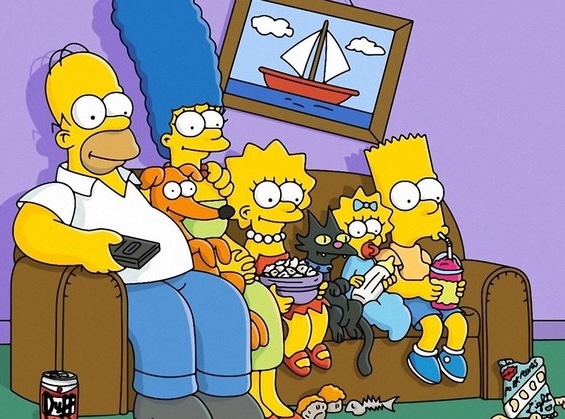}
        \caption{Source: United States}
        \label{fig:sub1}
    \end{subfigure}
    \hspace{0.03\columnwidth}
    \begin{subfigure}{0.45\columnwidth}
        \includegraphics[width=\linewidth,height=\linewidth]{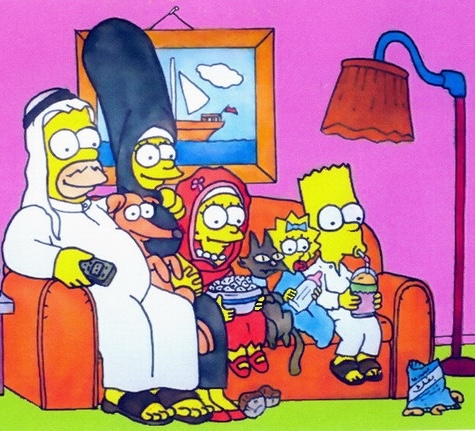}
        \caption{Target: Middle East}
        \label{fig:sub2}
    \end{subfigure}
    \caption{Image Transcreation in the real world: \textit{The Simpsons} (1989) was remade into \textit{Al-Shamshoon} (2005) for a Middle-Eastern audience.}
    \label{fig:simpsons_example}
\end{figure}

While MT has made commendable progress in the text and speech modalities \citep{kocmi2024preliminary, ahmad2024findings}, there has recently been interest in the ML community to approach translation in the \emph{visual} modality. This new task, called \textbf{image transcreation}, seeks to adapt (or localize) images originating from a \emph{source culture} for a given \emph{target culture} in an attempt to enhance familiarity and relatability for the target audience \citep{khanuja2024image}. We provide a popular real-world example of image transcreation in Figure \ref{fig:simpsons_example}, where content was recreated to match cultural norms of a target audience. While this problem is well-explored in translation studies (Section \ref{sec:related-work}) and is carried out widely in industrial settings by human translation experts (Section \ref{sec:applications}), it has not yet taken off in ML, despite the availability of powerful Vision-Language Models (VLMs). A central reason for this is the lack of automated evaluation mechanisms, with prior works relying solely on expensive human evaluation. \citet{khanuja2024image}, for instance, report spending about 4000 USD for annotating a test set of 3500 images across 7 countries, with only 1 annotator per image. While such a contribution is indeed beneficial, as ML researchers, we naturally need automated metrics to train, validate and benchmark image transcreation systems at scale.

And yet, the very nature of the task makes it difficult to propose a singular automated metric. Unlike traditional Natural Language Processing (NLP) problems like summarization or translation that have a relatively small space of valid outputs, there can be multiple valid ways of transcreating images. The suitability of an adaptation depends on factors like the intent/demands of the user/application, the creativity of the transcreator and the degree of cultural relevance required for the target audience. This subjectivity makes it hard to define a singular metric like BLEU \citep{papineni2002bleu}, ROUGE \citep{lin2004rouge}, or even neural metrics like COMET \citep{rei-etal-2020-comet} that optimize for similarity of model outputs against a fixed set of reference(s).

In this paper, we make progress towards building and evaluating automated metrics for image transcreation. First, given that this is a novel task in ML, we turn to translation studies for a better understanding of its scope and variables (Section \ref{sec:related-work}). We identify key dimensions critical to define the image transcreation process: \emph{cultural relevance}, \emph{semantic equivalence}, and \emph{visual similarity}, and look at real-world applications of transcreation through the lens of these dimensions in Section \ref{sec:applications}. Grounded in the conceptual and practical insights from previous sections, and drawing inspiration from MT metrics for text, we propose three metrics that account for the varied dimensions of this subjective task -- \textbf{a)} \textit{Object-based}; \textbf{b)} \textit{Embedding-based} and \textbf{c)} \textit{VLM-based} metrics, each of which leverages powerful open and closed VLMs and Large Language Models (LLMs). After conducting meta-evaluation along all three dimensions, we report average segment-level (i.e. instance-level) correlations with human ratings ranging from \textbf{0.55-0.87} on image transcreation test-sets spanning 7 countries. Thus, our primary contributions are:

\begin{enumerate}
    \item We propose three image transcreation metrics that are \textbf{a)} \textit{Object-based}; \textbf{b)} \textit{Embedding-based} and \textbf{c)} \textit{VLM-based} that achieve strong segment-level correlation, and to the best of our knowledge, are the first set of automatic evaluation metrics for this task.
    \item We assess the capabilities of powerful open and closed VLMs like Gemini, GPT-4o and Molmo to recognise, compare and reason over culturally specific objects in images, contributing to growing research in this area;
    \item Finally, we widen the scope of image transcreation for ML researchers by drawing from translation studies literature and elucidating complex real-world examples of this task. We hope this helps advance ML research on this challenging but important problem.
\end{enumerate}

\section{Transcreation in Translation Studies}
\label{sec:related-work}
Transcreation has been widely studied by linguists and philosophers in translation studies. We trace its development in this linguistics-adjacent field below, to ground our evaluation in this research.

\textbf{Definitions of Transcreation}: Transcreation, a concept introduced by the translation industry in the last century, has been developed to help clients of translation service providers tailor marketing campaigns for foreign markets \cite{ray2010reaching}, and is generally commonly associated with creative industries like advertising and entertainment. With its growing popularity, various definitions were proposed to underscore its difference from translation. Today, transcreation is generally described as a blend of linguistic translation, cultural adaptation, and creative reinterpretation of text elements \cite{diaz2023towards}. Given its ``cultural'' and ``creative'' aspects, it is also seen as a hybrid practice between ``translating'' and ``copywriting'' \cite{article}. As \citet{ray2010reaching} emphasize, transcreation focuses on effectively conveying the original message in a way that resonates with the target audience, even if this requires substantial deviations from the original text or imagery. The source text can be dethroned to apply focus on the target culture.


\textbf{Parameterizing Transcreation}: Broadly, as \citet{venuti2017translator} discusses, translation often navigates between the dichotomy of ``foreignization'', which emphasizes preserving the source culture, and ``domestication'', which adapts the content to fit the target culture. Various terms have been used to describe the ends of this spectrum, such as ``literal'' vs. ``free'', ``formal'' vs. ``dynamic'', and ``semantic'' vs. ``communicative''. Yet, no translation decision is ever solely one or the other \cite{venuti2017translator} and translators continuously navigate this spectrum, weighing several critical factors. 

In transcreation, this dichotomy is even more pronounced, and there is a strong emphasis on making the content feel relevant and engaging in a new cultural context, especially in areas like marketing and entertainment. This often requires modifying or replacing Culturally Specific Items (CSIs) when source-cultural references do not easily transfer to the target audience \cite{aixela1996culture}. The goal is not just to ensure cultural relevance but also to maintain the intended impact of the original content. Thus, every transcreator must strike a balance between preserving \emph{semantic equivalence} -- ensuring the intended meaning remains intact, and achieving \emph{cultural relevance} by adapting the content for the target audience. 

A crucial factor in determining this balance is the intent behind the transcreation, as emphasized by Skopos theory \cite{Reiss2014}. The intent can be as broad as the overall goals of a marketing campaign or as specific as the instructions in a project brief \cite{diaz2023towards}. Whether the purpose is to inform, persuade, or entertain, this intent dictates the degree to which the original content should be adapted \cite{ray2010reaching}.

Finally, transcreation is inherently a creative process. Beyond fidelity to the source, it demands an imaginative reworking of content to achieve cultural resonance \cite{marian2024}. ``Creativity'' is the mechanism through which cultural relevance can be achieved. As \citet{Nord2009a} points out, this creative process often demands unique, non-replicable solutions tailored to the specific needs of each project. While creativity itself may not be easily quantifiable, evaluating \emph{visual similarity} between the original and transcreated content provides an indirect means of assessing how faithfully the adaptation retains key visual elements while achieving its intended cultural and semantic transformation.

\section{Visual Transcreation in the Wild}
\label{sec:applications}
While text translation operates in a relatively constrained and predictable output space, image transcreation allows for a broader range of creative expression and the desired output depends on the user’s objectives and application context, as elucidated above. This \textbf{intent} or purpose of transcreation also drives the extent of (a) \textit{cultural relevance}, (b) \textit{semantic equivalence}, and (c) \textit{visual similarity} needed in the final output, as we describe below: 

\begin{figure}[!t]
    \centering
    \includegraphics[width=\linewidth]{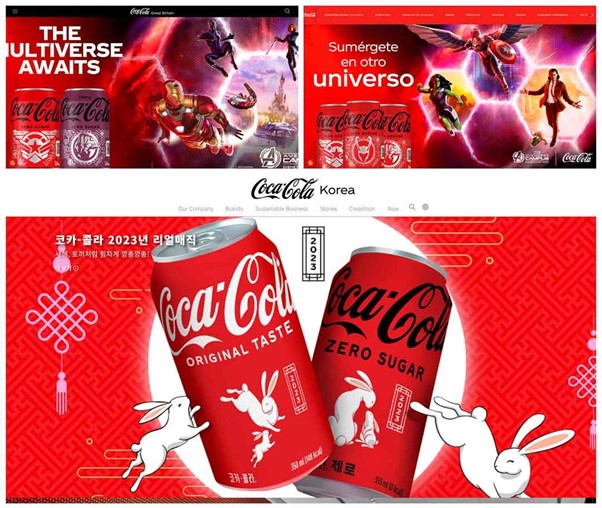}
    \caption{The Coca-Cola homepage depicts different cultural concepts in different countries \citep{cocacolaad}. The top row shows Marvel-themed ads where visuals are consistent but the text is translated for accessibility, while the bottom one depicts the Lunar New Year to celebrate the Year of the Rabbit in Korea.}
    \label{fig:cocacolaad}
\end{figure}

\begin{figure*}[h]
    \centering
    \begin{subfigure}{0.45\textwidth}
        \centering
        \includegraphics[width=\textwidth]{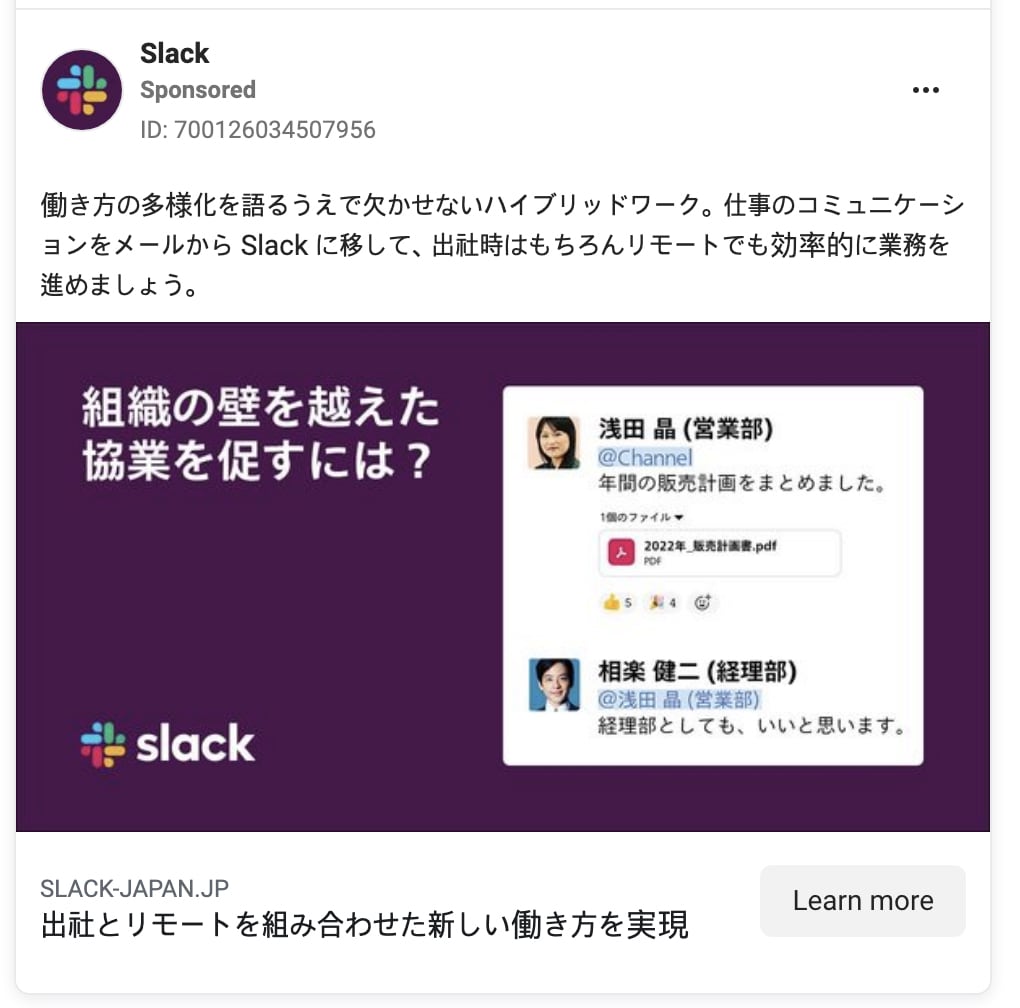}
        \caption{Japan advertisement}
        \label{fig:slackjpn}
    \end{subfigure}\hfill
    \begin{subfigure}{0.45\textwidth}
        \centering
        \includegraphics[width=\textwidth]{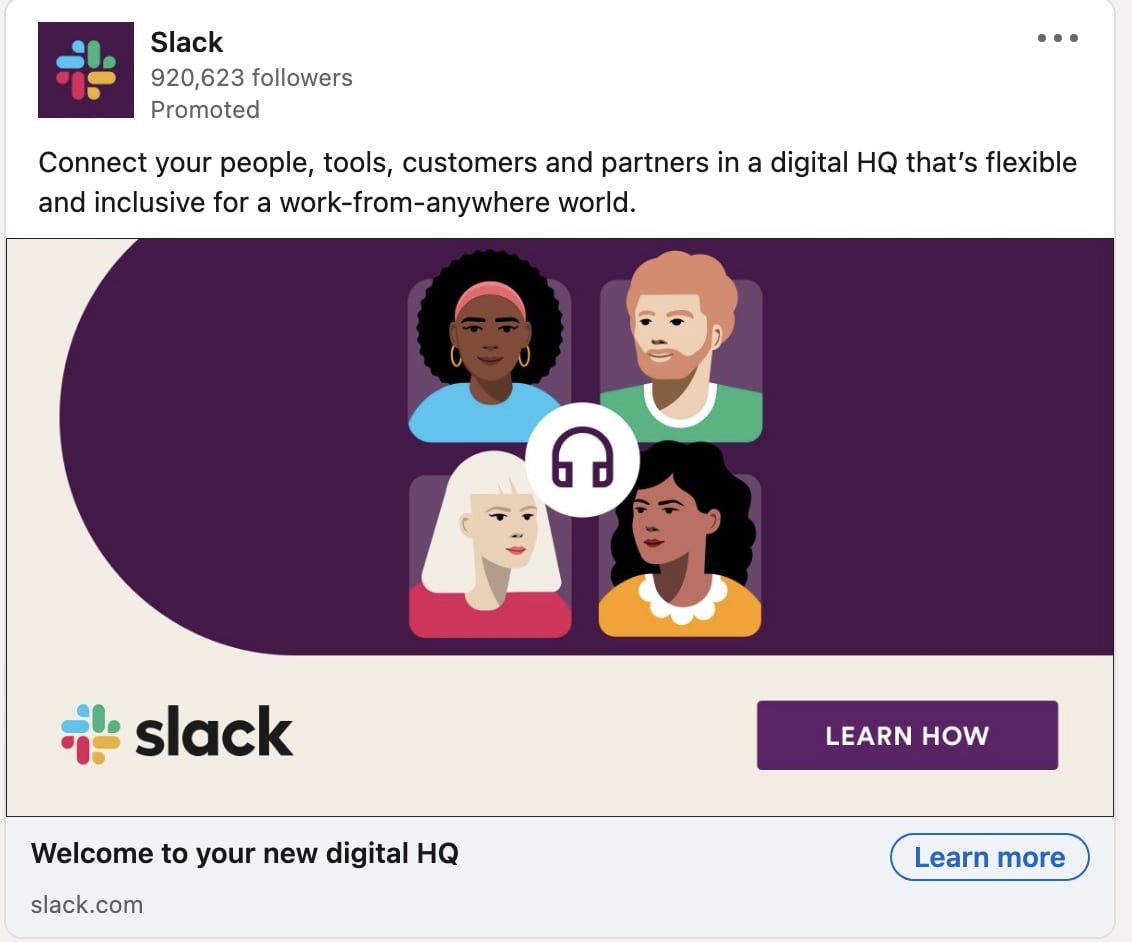}
        \caption{US advertisement}
        \label{fig:slackus}
    \end{subfigure}
    \caption{Both Slack ads convey how the product can be used to facilitate hybrid work, but the names and avatars of users are localized to reflect the demographics of each country. Figures sourced from \citet{slackblog}. }
    \label{fig:slack}
\end{figure*}

\subsection{Advertisements}
\label{sec:ads}
\noindent\textbf{Intent}: \emph{to emotionally relate to the target audience, while preserving the core brand identity}

Given the strong focus on being relatable to the target audience to maximize product sales, we often see ad transcreations span a broad spectrum of creativity. A well-known example is Coca-Cola,
which embodies its slogan, ``Think Global, Act Local'', by maintaining brand consistency while seamlessly integrating cultural elements into its ads. For example, they released a Lunar New Year ad in Korea to celebrate the year of the rabbit (Figure \ref{fig:cocacolaad}). In a particularly ambitious example, they transcreated an entire video ad, using different actors speaking in different languages, but with the script being exactly the same otherwise \citep{cocacola2023}. Beyond region-specific adaptations, some ads also focus on universal themes while still targeting specific audiences, such as their Marvel-themed campaign (Figure \ref{fig:cocacolaad}). When released in different regions, these are minimally transcreated, often only translating the text while keeping the visuals unchanged. In some cases, subtle visual differences can emerge across regions, even when ads present similar concepts. For instance, in Figure \ref{fig:slack}, both Slack ads promote hybrid work, but the Japanese and U.S. versions modify the names and avatars to better reflect the demographics of each country.

Overall, transcreators exhibit a strong focus on \textit{cultural relevance} of the final output, and preserving the exact meaning of the source image and even its spatial layout may not be necessary (as we see in the Marvel / Lunar New-Year example). Hence, transcreators would prefer models that can make outputs highly \textit{culturally relevant}, while \textit{semantic equivalence} and \textit{visual similarity} can be given lower priorities.

\subsection{Comics}
\label{sec:comics}
\noindent\textbf{Intent}: \emph{to adapt the story and characters in a way that remains faithful to the original while making them accessible to a new audience}

In the early days of comic and manga transcreation, adaptations often prioritized cultural relevance and accessibility for the target audience. Translators and publishers frequently made significant changes to align with local norms and expectations, believing this would enhance the appeal of the material. However, over time, audiences prefer content that remains faithful to its original form, valuing authenticity over localized adjustments.  This is best exemplified by the evolution of Japanese manga transcreation for Western audiences. In early days, U.S. publishers believed that altering key elements like reading direction would make it more accessible. In \textit{Akira}, for instance, pages were flipped for left-to-right reading, but this caused significant cultural errors. Japanese cars were incorrectly depicted as driving on the right, and characters were shown left-handed, even in situations where this affected key plot points, such as losing their dominant hand in a fight \citep{gianordoli2023manga}. Over time, it became clear that \textbf{fidelity is strongly preferred over creativity} in the transcreation of comics, particularly by manga fans.\footnote{\url{https://tinyurl.com/58vzy6xu}}\textsuperscript{,}\footnote{\url{https://tinyurl.com/5c6wua42}}

\begin{figure*}[h]
    \centering
    \begin{subfigure}{0.5\textwidth}
        \centering
        \includegraphics[width=\textwidth]{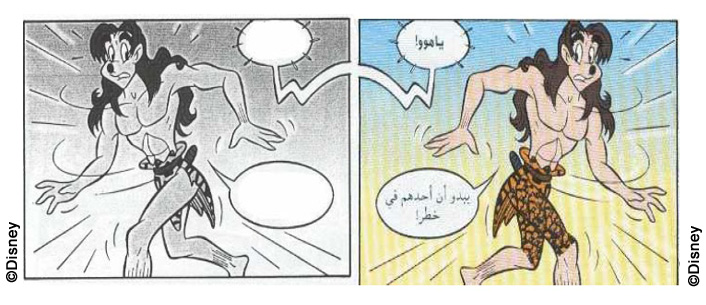}
        \caption{Brad of the Jungle (Kuwaiti version)}
        \label{fig:tarzankw}
    \end{subfigure}\hfill
    \begin{subfigure}{0.5\textwidth}
        \centering
        \includegraphics[width=\textwidth]{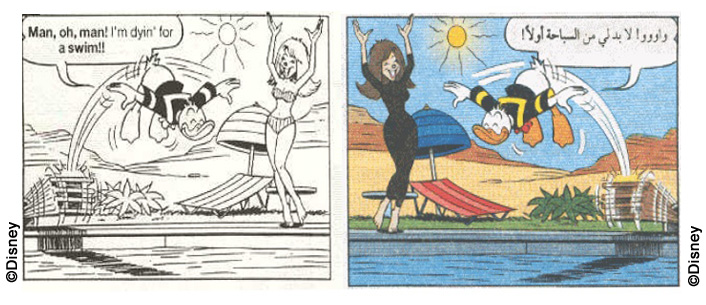}
        \caption{Donald Duck (UAE adaptation)}
        \label{fig:dduae}
    \end{subfigure}
    \caption {Examples of transcreation to adhere to cultural norms. \emph{Left}: Tarzan covers his thighs with culturally appropriate clothing; \emph{Right}: woman is completely dressed in black.}
    \label{fig:me_adaptations}
\end{figure*}

\begin{figure}[h]
        \centering
        \includegraphics[width=\linewidth]{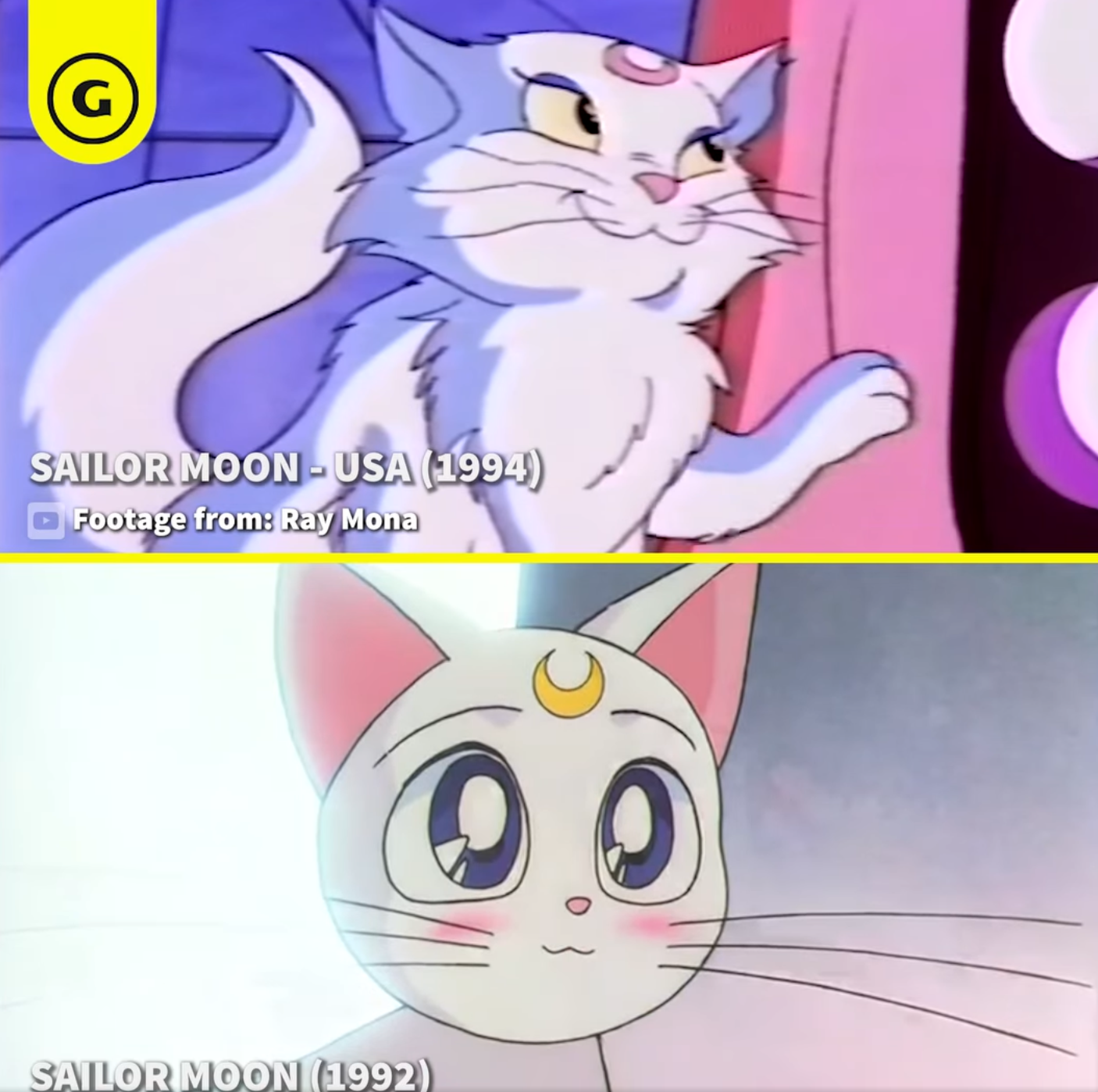}
        \caption{\emph{Transcreation reflecting cultural differences} -- in the U.S. adaptation of \textit{Sailor Moon}, \textit{Artemis} is portrayed as cunning and mischievous (``catty"), while in Japan, cats are seen as part of the cute \textit{kawaii} culture.}
        \label{fig:sailormoon_cat_diff}
\end{figure}

However, there are still instances where adaptation is unavoidable, often driven by the need to align with social norms, legal requirements, cultural differences, or when concepts are completely absent in the target culture. For example, when Disney comics like \textit{Tarzan} and \textit{Donald Duck} were adapted for Kuwait and the UAE, changes were made to clothing to conform to local cultural expectations (Figure \ref{fig:me_adaptations}). Similarly, in Marvel’s \textit{The Punisher}, a swastika tattoo on a villain was replaced with a parallelogram when adapted for Germany, adhering to the country’s strict legal restrictions on the depiction of Nazi symbols \citep{zanettin2014visual}. In the US TV adaptation of the manga, \emph{Sailor Moon}, the character of \textit{Artemis}, a cat, was creatively adapted to match the stereotypes and perceptions of American culture (Figure \ref{fig:sailormoon_cat_diff}) -- showing transcreation also needs to factor in cultural ideas and beliefs.
Finally, we note that Japanese manga often uses onomatopoeia to represent sounds, which is less common in Western comics. For instance, in Tsuge Yoshiharu’s \textit{The Swamp}, the sound of a rifle crack was changed from “zdom” to “kbooom, a term more familiar for the target audience \citep{gianordoli2023manga}.


Overall, we note that while readers typically prefer transcreations that remain true to the original, adaptations are sometimes necessary to meet social norms, legalities, or cultural differences. Hence, models that pay more attention to \textit{semantic equivalence} and \textit{visual similarity} would be preferred, even at the cost of lower overall \textit{cultural relevance} for the target audience.

\subsection{Children's Media}
\label{sec:kidsmedia}
\noindent\textbf{Intent}: \textit{to aid learning and cognitive development through familiar objects, concepts and experiences}

\begin{figure}[h]
        \centering
        \includegraphics[width=\linewidth]{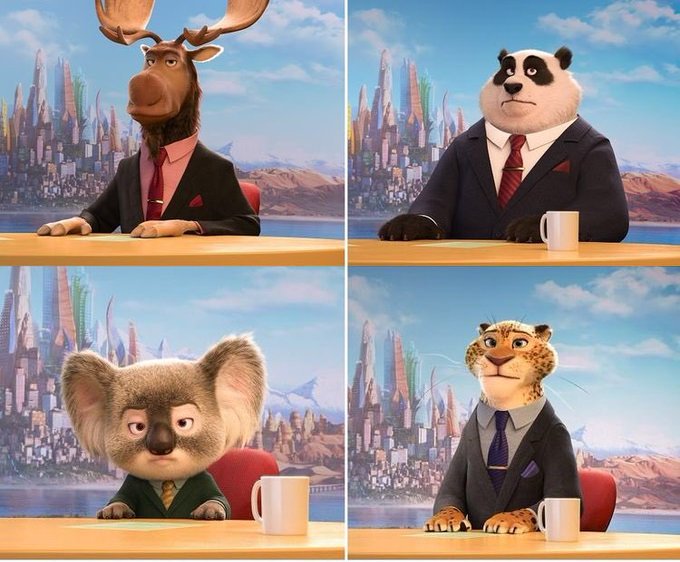}
        \caption{The newscaster in the Pixar movie "Zootopia" was transcreated based on the country: a \textit{moose} for the US/Canada/France, a \textit{panda} for China, a \textit{koala} for New Zealand/Australia, and a \textit{jaguar} for Brazil.}
        \label{fig:zootopia_transcreation}
\end{figure}

Several studies highlight how incorporating familiar imagery and references improves cognitive engagement for children by making abstract concepts easier to grasp \cite{unicef2018learning,purcell2011building}. We see several examples of this, for example in Pixar and Disney movies, like in \textit{Zootopia} (2016) where the character of the newscaster was adapted to match a country's national animal (Figure \ref{fig:zootopia_transcreation}). Another prominent example is \emph{Dora the Explorer}, where Dora switches between Hindi and English in its adaptation for India, references local animals like the tiger and peacock, and also talks about issues affecting female children like early marriage, and unequal distribution of household chores \citep{khalid2012meena}. This approach not only promotes bilingual learning and concept-learning, but also empowers children to challenge social norms. \emph{Sesame Square}, Sesame Street's adaptation in Nigeria, similarly had Muppets who spoke with a Nigerian English accent and tackled weighty issues such as HIV/AIDS discrimination, hygienic practices and ways to cope with death \citep{sangillo2019sesame}. Beyond movies and TV, educational apps like Toca Boca also incorporate region-specific foods (like sushi in Japan), festivals (like the Lunar New Year in China), and clothing styles, to ensure that children are interacting with culturally familiar elements while learning new concepts.



Overall, we again notice a stronger focus on \textit{cultural relevance} for the target audience, which can come at the cost of \textit{visual similarity} or complete \emph{semantic equivalence}, even though themes are similar (like in the adaptation of Dora the Explorer or Spiderman for India). However, in certain cases like for Pixar movies, which specifically focus on creating global content, minimal changes are made for \textit{cultural relevance}, and \textit{visual similarity} + \textit{semantic equivalence} takes precedence.

\section{Proposing Automatic Metrics}
\label{sec:evaluation}
\citet{khanuja2024image} conduct a multi-faceted human evaluation to assess image transcreation systems. Despite being a great first step, a key limitation of the human evaluation questions they design is the lack of grounding in related work from adjacent fields, as well as their specificity to the test set they create. From the theories we study in Section \ref{sec:related-work}, and by diving deep into what applications actually need (Section \ref{sec:applications}), we identify three factors critical to a transcreation process, each of which may have different preferences based on the user/task: namely \emph{cultural relevance}, \textit{semantic equivalence}, and \emph{visual similarity}. Hence, we must design metrics that can capture all three dimensions.

But how do we design such metrics? Similar to text-level translation that aims to transfer source-side semantics to a target language, during image transcreation, the goal is to recreate visual ideas for a target culture. Tracing the history of MT evaluation, we observe that it first started with human annotators evaluating the adequacy, fluency, and overall quality of the translation. Given high costs and time latency associated with this process, automatic metrics began to develop, which can be broadly categorized into \emph{lexical-based}, \emph{embedding-based}, and more recently, \emph{LLM-based} metrics \cite{fu2023gptscore,liu2023g}. Lexical-based metrics measure similarity between the hypothesis and reference by a mathematical formula, and are word-based or character-based depending on the matching units (BLEU \citep{papineni2002bleu}, METEOR \citep{meteor}). Embedding-based metrics use dense vector representations of hypotheses and references to measure their similarity \cite{BERTScore, bartscore,lo2019yisi}, and metrics like COMET \citep{rei-etal-2020-comet} also train regression models on human judgments of translation quality. Finally, LLM-based metrics have recently emerged, leveraging LLMs to evaluate translations by using their contextual understanding and reasoning capabilities. For example, GPTScore \citep{fu2023gptscore} showed how GPT could be prompted to score hypotheses. Further, G-Eval \citep{liu2023g} leveraged GPT-4 for evaluating generation based on open-ended reasoning, while Gemba-MQM \cite{kocmi-federmann-2023-gemba} used the same model to detect translation error spans and achieved SOTA system-level correlation in the WMT'23 Metrics Shared Task. 



Drawing from the above lineage, a parallel evolution of metrics for image transcreation is conceptualised below: 

\subsection{Object-based metrics} 
\label{sec:object}
Inspired by BLEU \cite{papineni2002bleu}, we propose an object-based metric that evaluates how well the model replaces culturally-salient objects with valid replacements, given the intent and target demographic. Concretely, this would involve:
\begin{itemize}
    \item \textbf{Step 1: Object Identification}: Identify all objects in both the source and target (model output) images.

    \item \textbf{Step 2: Obtain Valid Translations}: Conditioned on the intent of transcreation, identify the culturally-specific items (CSIs) \cite{aixela1996culture} in the source image, that would be unfamiliar to the target culture. Subsequently, determine a valid set of replacements for each CSI based on the constraints of this intent.
    
    \item \textbf{Step 3: Metric Calculation}: Calculate the proportion of CSIs that have been correctly replaced in the model output. This is determined by checking whether objects identified in the target image are valid replacements of the source CSIs according to \emph{Step 2}.
    
\end{itemize}





For example, let's say we are transcreating a US-based image for Japan, and the source image contains three objects identified as CSIs: broccoli, spoon, and a dollar bill. Let's say that given a generic intent of \emph{relatability to the target audience}, valid replacements are identified as: \textbf{a)} broccoli: \{bell pepper, radish\}; \textbf{b)} dollar bill: \{yen note\}; \textbf{c)} spoon: \{chopsticks\}.

Now, consider the model output contains the following objects: bell pepper (correct replacement for broccoli), spoon (not replaced) and a euro bill (incorrect replacement for dollar bill). Given the model correctly replaces one out of three CSIs, we get a score of 0.33.
\[
\text{CSI-Overlap} = \frac{1 + 0 + 0}{3} = \frac{1}{3} \approx 0.33
\]

\subsection{Embedding-based Metrics}  
In MT, embedding-based metrics compare contextual embeddings of the target and reference texts. For image transcreation, collecting large amounts of parallel data seems infeasible at present. Instead, we can leverage powerful vision-language encoders to evaluate transcreation outputs across the three axes discussed in Section \ref{sec:applications}:

\noindent \textbf{a)} \emph{cultural relevance}: can be measured by evaluating the change in similarity between the visual embeddings of the source \( V_{\text{source}} \) and transcreated image \( V_{\text{target}} \) relative to a cultural reference text \( T_{\text{culture}} \), here, "This image is from region \texttt{X}." Mathematically:

\[
\Delta S_{\text{c-r}} = \text{cosine}(V_{\text{tgt}}, T_{\text{cul}}) - \text{cosine}(V_{\text{src}}, T_{\text{cul}})
\]

\noindent where \( \Delta S_{\text{c-r}} \) represents the change in similarity that indicates cultural relevance. A higher increase in \( \Delta S_{\text{c-r}} \) indicates a greater change towards cultural relevance for the target demographic. \\


\noindent \textbf{b)} \emph{semantic equivalence}: can be measured by the similarity between the transcreated image's visual representation \( V_{\text{target}} \) and the textual representation of the intent \( T_{\text{int}} \). For example, if we are transcreating one food item into another food item, this can be measured using the cosine similarity between the target image embeddings and the text, "This is a food item". Mathematically:

\[
S_{\text{s-e}} = \text{cosine}(V_{\text{target}}, T_{\text{int}})
\] 

\noindent where \( S_{\text{s-e}} \) represents the semantic equivalence between the transcreated image and the intent of transcreation. \( V_{\text{target}} \) is the visual embedding of the transcreated (target) image, and \( T_{\text{int}} \) is the textual representation of the intent. A higher \( S_{\text{s-e}} \) value indicates a closer alignment with the intended semantic meaning. \\

\noindent \textbf{c)} \emph{visual similarity}: can be measured by the similarity between visual representations of the source and transcreated images. A high similarity indicates that the image has stayed close to the source, while a lower similarity suggests a greater degree of adaptation. Mathematically:

\[
S_{\text{v-s}} = \text{cosine}(V_{\text{src}}, V_{\text{tgt}})
\]

\noindent where \( S_{\text{v-s}} \) represents the visual similarity between the source and transcreated (target) images. \( V_{\text{src}} \) and \( V_{\text{tgt}} \) are the visual embeddings of the source image and the transcreated image, respectively.




\subsection{VLM-based Metrics} 
Finally, similar to LLM-based scores for text, we can prompt VLMs to assign a score on a fixed scale to estimate how far the model has transcreated the source image for \emph{cultural relevance}; how \emph{semantically equivalent} is the transcreated image to the source image; and how they fare on \emph{visual similarity}. Prompt formats can leverage chain-of-thought reasoning \cite{liu2023g}, and our implementation of the same is discussed in Section \ref{sec:correlation}.

\section{Experimental Setup}
\label{sec:experiments}
In this section, we perform a meta-evaluation of the transcreation metrics introduced above. Specifically, we assess the outputs of transcreation systems proposed in \citet{khanuja2024image} using the metrics defined in Section \ref{sec:evaluation} and analyze the correlation between our scores and human ratings for the same system outputs.\footnote{Available at: \url{https://github.com/simran-khanuja/image-transcreation/tree/main/human_evaluation}}
\subsection{Evaluation Dataset}

\begin{table}[ht]
\centering
\renewcommand{\arraystretch}{0.9} 
\setlength{\tabcolsep}{3pt} 
\resizebox{\columnwidth}{!}{
\begin{tabular}{lclclclc}
\toprule
Country & Images & Country & Images & Country & Images & Country & Images \\
\midrule
Brazil & 501 & India & 504 & Japan & 509 & Nigeria & 506 \\
Portugal & 506 & Turkey & 509 & USA & 508 & Total & 3543 \\
\bottomrule
\end{tabular}
}
\caption{Dataset size for the 7 target countries.}
\label{tab:eval_stats}
\end{table}



\noindent \citet{khanuja2024image} construct a dataset featuring commonly occurring concepts across 17 semantic categories (such as food, beverages, and mammals) that are culturally comprehensive, with data collected from seven different countries. The final source dataset consists of 585 images. Each of the seven countries serves as a target, with source images drawn from the remaining six countries, resulting in approximately 500 images being transcreated for each target country (Table \ref{tab:eval_stats}). Further details on this dataset can be found here.\footnote{\url{https://machine-transcreation.github.io/image-transcreation/}} The task involves converting an image into another within the same semantic category while ensuring greater cultural relevance for a specific target country, X. For example, a food item in the source image may be replaced with a different food item that is more commonly recognized or significant in X.

\subsection{System Descriptions}

\citet{khanuja2024image} propose three systems comprising of state-of-the-art LLMs, retrievers and image-editing models for the image transcreation task. Briefly, the first system, \textbf{\texttt{e2e-instruct}}, is a trivial end-to-end image-editing model prompted to \emph{make the image culturally relevant to X}, where X is the target country. The second pipeline, \textbf{\texttt{cap-edit}}, first captions the image, then makes the caption `culturally relevant' using a LLM, and finally edits the original image with the culturally-modified caption. The third system, \textbf{\texttt{cap-retrieve}} uses the culturally-modified caption from \texttt{cap-edit} to retrieve a natural image using the modified caption as the query. In their dataset, all outputs are given a rating on a scale of 1-5 by human annotators, across multiple dimensions including visual change, semantic equivalence, cultural relevance, naturalness and offensiveness. They find that the \texttt{cap-retrieve} pipeline performs best, across countries. Questions relevant to our meta-evaluation are in Table \ref{tab:qs}. 

\begin{table}
\centering
\small 
\begin{tabular}{p{5.4cm} p{1.7cm}}
\toprule
\textbf{Questions from \cite{khanuja2024image}} & \textbf{Category} \\ 
\midrule
Does the image seem like it came from your country / is representative of your culture? & \texttt{cultural relevance} \\
\addlinespace
Is the generated image from the same semantic category as the original image? & \texttt{semantic equivalence} \\
\addlinespace
Is there any visual change in the generated image compared to the original image? & \texttt{visual similarity} \\ 
\addlinespace
\bottomrule
\end{tabular}
\caption{Questions from the evaluation set of \citet{khanuja2024image} used for meta-evaluation in this work.}
\label{tab:qs}
\end{table}


\subsection{Metric Implementation}
\label{sec:correlation}
We calculate segment-level correlations of proposed metric scores with human ratings. Similar to MT that defines segment-level correlation on the order of test instances (sentences), we define it at the level of each image in our test set. We use the default SciPy (v1.11.4) implementation of Kendall's Tau \citep{kendall1938new}, and average across the entire test set for a given country. Segments with non-parsable model outputs, or those scored identically by all systems (usually a small minority) are skipped. Our metrics are implemented as follows:

\textbf{Object-based (CSI-Overlap)}: For the first step of object identification, we prompt Gemini-1.5-Pro \citep{reid2024gemini}, since current open VLMs have been shown to perform poorly on open-vocabulary, long-tailed object detection \cite{gupta2019lvis}. For the second and third steps of obtaining valid replacements and calculating the proportion of CSIs correctly replaced, we again leverage Gemini-1.5-Pro, but any LLM can be used for this purpose. 

\textbf{Embedding-based}: We use SigLIP \cite{zhai2023sigmoid}, a dual-encoder VLM that improves on CLIP \citep{radford2021learning} by using a sigmoid loss instead of softmax, allowing for stronger alignment between images and text without large-scale resources.  To calculate \emph{cultural-relevance}, we obtain text embeddings for ``This image belongs to country \texttt{X}", where \texttt{X} denotes the target country for transcreation. To estimate \emph{semantic equivalence}, we similarly obtain text embeddings of the statement ``This image belongs to category \texttt{X}", where \texttt{X} is the semantic category the image belongs to, like `food', `beverage', `mammal' etc.

\textbf{VLM-based}: Finally, we prompt the closed VLMs Gemini-1.5-Pro, GPT-4o \citep{OpenAI_HelloGPT4o}, and the open-sourced Molmo 7B-D model \citep{deitke2024molmo} to obtain VLM-based scores (prompts listed in Appendix \ref{sec:appendix}). 



\subsection{Results}

Table \ref{tab:correlation-results} presents the average segment-level correlations across seven countries for the three evaluation criteria. 

For \emph{semantic equivalence}, proprietary VLMs such as GPT-4o and Gemini-1.5-Pro show the highest performance, likely due to their advanced multimodal reasoning capabilities. We also note that embedding-based metrics outperform open-source VLMs, thus proving to be more cost-effective in budget-lean scenarios. However, as open-source VLMs improve, these dynamics could shift in favor of open-source VLMs. 

For \emph{visual similarity}, embedding-based metrics using the SigLIP model excel, achieving a strong correlation of 0.87, significantly surpassing both open-source and proprietary VLMs. This suggests that embedding models are better at capturing fine-grained visual nuances than VLMs that primarily focus on image-text alignment during training. 

For \emph{cultural relevance}, the trends mirror those of semantic equivalence, though with lower overall correlations, highlighting the complexity of assessing cultural relevance and the potential for future research in building metrics that capture this dimension.

\begin{table}
\centering
\resizebox{\columnwidth}{!}{%
\begin{tabular}{l@{\hskip 8pt} l@{\hskip 8pt} c c c}
\toprule
\textbf{Category} & \textbf{Model} & \textbf{cult-rel} & \textbf{sem-eq} & \textbf{vis-sim} \\ 
\midrule
\textbf{Object-based} & Gemini-1.5-Pro & -0.08 & 0.29  & -0.16 \\ 
\addlinespace
\cdashline{1-5}
\addlinespace
\textbf{Embedding-based} & SigLIP & 0.35 & 0.58 & \textbf{0.87} \\ 
\addlinespace
\cdashline{1-5}
\addlinespace
\multirow{3}{*}{\textbf{VLM-based}} & Gemini-1.5-Pro & \textbf{0.81}  & 0.48 & 0.52 \\ 
 & GPT-4o & 0.55 & \textbf{0.86} & 0.17   \\ 
 & Molmo 7B & -0.16 & 0.44 & 0.12  \\ 
\addlinespace
\bottomrule
\end{tabular}
}
\caption{Average segment-level correlation between human ratings and the three automated metrics proposed. `cult-rel' -- \texttt{cultural relevance}, `sem-eq' -- \texttt{semantic equivalence}, `vis-sim' -- \texttt{visual similarity}. }
\label{tab:correlation-results}
\end{table}

\subsection{Strengths, Limitations, Recommendations}
We are inspired by recent research on meta-evaluation in MT that has shown that no single MT metric is effective in capturing all kinds of translation errors and each has its strengths and weaknesses, making a singular overall score from a `winning' metric uninformative \citep{moghe2024aces}. We argue that this problem is exacerbated in transcreation, where the open-ended and application-dependent nature of the problem might make an overall score ambiguous. Thus, we now analyze the strengths and limitations of each metric and provide recommendations for the metric models that researchers might use in different scenarios. Section \ref{sec:applications} is an attempt at elucidating the axes of transcreation which can potentially be more or less relevant for a given application. We also qualitatively analyze the differences across categories in the transcreation dataset that contains 17 categories ranging from concrete (e.g., fruits, birds, beverages), to abstract (e.g. education, religion) ones. Category-wise correlation scores with human ratings can be found here.\footnote{\url{https://github.com/simran-khanuja/automatic-eval-img-transcreation/tree/main/category-correlations}} 

\textbf{Object-based} metrics share some limitations with metrics like BLEU. It focuses on exact object matches and offers no credit for near-correct adaptations. Furthermore, like BLEU’s reliance on n-gram overlap, this metric does not account for the context or positioning of objects, and hence cannot capture awkward or nonsensical edits. In general, while it captures \emph{cultural relevance} and \emph{semantic equivalence} somewhat, it fails to compute the extent of \emph{visual similarity} in adaptation. Looking at category-wise correlation scores for \emph{cultural relevance} and \emph{semantic equivalence}, it performs well with concrete categories (e.g., beverages, vegetables, fruits, mammals), but struggles with abstract ones (e.g., celebrations, religion, education).
Overall, it obtains the lowest correlation values across all three axes, likely due to its modular approach facilitating error propagation.

\textbf{Embedding-based} metrics that leverage vision-language encoder representations excel at capturing visual similarity, even more so than proprietary SOTA VLMs. Gains are especially notable for concrete categories, though slightly diminished for abstract ones (e.g., houses, education, religion). While they are not as competent in multimodal reasoning, they offer a cost-efficient alternative to evaluate systems across all three dimensions. 

\textbf{VLM-based} metrics are highly effective in capturing \emph{cultural relevance} and \emph{semantic equivalence} due to their advanced multimodal reasoning abilities. For \emph{cultural relevance}, they perform particularly well for culturally significant categories (e.g., celebrations, religion, music, visual arts), but less so
for generic ones (e.g., flowers, fruits, birds). For instance, VLMs rate culturally unique concepts like \emph{Holi} highly but assign neutral scores to more generic objects like \emph{roses} -- despite the latter holding cultural significance in certain contexts, like in India. For \emph{semantic equivalence}, VLMs handle concrete categories (e.g. food, beverages, mammals) effectively but, like CSI-Overlap, struggle with abstract ones (e.g. religion, education). Lastly, in terms of visual similarity, VLMs are weaker than vision encoders, though they perform better with concrete categories (e.g., birds, beverages, clothing) than abstract ones. 

We conclude by noting that despite the relatively higher scores, even proprietary VLMs exhibit only moderate absolute correlations with human ratings on cultural relevance, underscoring the complexity of this task and leaving headroom for improvement.

\textbf{Overall}, our findings suggest that a hybrid approach---leveraging proprietary VLMs for reasoning-heavy tasks like \emph{cultural relevance} and \emph{semantic equivalence} and embedding models for \emph{visual similarity}, constitutes the most comprehensive evaluation strategy. We also note that the evaluation of abstract cultural categories remains a challenging and open problem for future work -- although VLMs do show some promise in this regard.

\section{Conclusion}
\label{sec:conclusion}

In this paper, we take a first step towards automatically evaluating image transcreation - the task of adapting visual content from a source culture to a target culture. We first review related works in translation studies and identify that \emph{cultural relevance} for the target audience, \emph{semantic equivalence} of the transcreated image, and its \emph{visual similarity} with the source image are three measurable dimensions which potentially cover a wide range of applications. We explore how visual content is being transcreated for real-world applications through this lens, and design evaluation metrics that rank systems across these three axes, drawing inspiration from MT metrics. We find that VLM-based metrics excel at measuring \emph{cultural relevance} and \emph{semantic equivalence}, but visual similarities can be better captured using visual-encoder embeddings. For all metrics, capturing cultural relevance of an image for a target demographic is challenging, leaving headroom for progress. We hope that our study helps ML researchers make progress on this important, real-world task. 

\section{Limitations}
Our work does not come without limitations. First, even though we broaden the scope of image transcreation to several applications and study real-world examples, we do not cover examples from videos and gaming, the latter being one of the largest markets for localization. However, this is a very ambitious scope and we hope to slowly make progress toward the same. Second, capturing both cultural relevance and semantic equivalence can be very nuanced. We rely on country or region names to capture the former, and textual/category descriptions for the latter, both of which do not capture finer-grained nuances across systems. Thirdly, since this is a new task, we only have one evaluation dataset available \cite{khanuja2024image} -- which only provides scores for three systems, with only one annotator rating per target image. \textbf{\texttt{Cap-retrieve}} is overwhelmingly better than the other two systems which might make ranking these systems a relatively easier task. We hope that as the community develops more image transcreation models, we should be able to test whether VLMs can effectively capture nuanced differences between systems as well. Lastly, all our prompting for VLM-based metrics is in English, but it would be quite interesting to explore extensions to other languages, particularly those native to the target culture, and we leave this for future work.

\section{Acknowledgements}
We were fortunate to obtain feedback on this work from a variety of sources. In particular, we would like to thank Swaminathan Gurumurthy and members of Neulab at Carnegie Mellon University, as well as Barry Haddow, Alexandra Birch, and the StatMT group at the University of Edinburgh. This work was supported in part by grants
from Google and the Defense Science and Technology Agency Singapore. Computations in this work were facilitated by the Baskerville Tier 2 HPC service (\url{https://www.baskerville.ac.uk}). Baskerville was funded by the EPSRC and UKRI through the World Class Labs scheme (EP/T022221/1) and the Digital Research Infrastructure programme (EP/W032244/1) and is operated by Advanced Research Computing at the University of Birmingham.


\section{Ethical Statement}


In this work, we emphasize that we are not attempting to formalize the creation of cultural deepfakes or undermine the jobs of artists. On the contrary, the goal is to increase accessibility to culture-specific content. We recognize that the demand for human-generated translations and human creativity will always be paramount.

At the same time, we acknowledge that some people prefer to engage directly with the source culture rather than consume localized adaptations. This is particularly true among manga fans, who often take pride in understanding nuanced cultural references that may be lost in translation. In such fan communities, translations and localizations are frequently criticized, as they inevitably result in a distortion or transformation of the original author's intended meaning. While localized adaptations can expand the global audience, we believe there is inherent value in respecting cultures in their unadulterated form and providing opportunities for people to engage directly with source materials. Our aim is to strike a balance between accessibility and preserving the integrity of cultural works.

\section{Contributions}
We list author contributions loosely following the CRediT author statement.\footnote{\url{https://www.elsevier.com/researcher/author/policies-and-guidelines/credit-author-statement}}

\begin{itemize}
    \item \textbf{SK}: Conceptualization (co-lead), Methodology (lead), Software (co-lead), Resources (lead), Writing (co-lead), Investigation (co-lead), Visualization (co-lead)
    \item \textbf{VI}: Conceptualization (co-lead), Software (co-lead), Writing (co-lead), Investigation (co-lead), Visualization (co-lead)
    \item \textbf{CH}: Writing (supporting), Investigation (supporting), Resources (supporting)
    \item \textbf{GN}: Supervision (lead), Project Administration (lead), Funding Acquisition (lead), Writing (supporting)
\end{itemize}

\bibliography{anthology,custom}
\bibliographystyle{acl_natbib}

\appendix

\section{Appendix}
\label{sec:appendix}

All the prompts used to calculate the CSI-Overlap metric can be found in Table \ref{tab:prompts} below.

\onecolumn
\begin{tcolorbox}[colback=white,colframe=black,boxrule=0.5mm,arc=4mm,title=Prompts used to calculate CSI Overlap in Section 4.1]
\ttfamily 
\textbf{Gemini-1.5 Pro (Object Detection)} \\
"You are an expert in detecting objects in an image. Detect unique objects/entities in the image and only list each object ONCE. Give the Wikipedia or Wikidata entity name of the object detected, if possible. The output format is a comma-separated list of entities." \\

\textbf{Gemini-1.5-Pro (Getting Replacements)} \\
"Imagine you are an image editor tasked with adapting visuals for a global audience. You are given an image where object detection identified these entities: {', '.join(extracted\_objects)}. Your goal is to identify any entities that represent specific countries, cultures, or regions and are not be universally recognizable or relevant to someone from \texttt{COUNTRY}. For each culture-specific entity you identify, please provide a diverse and exhaustive list of potential replacements that would be more culturally relevant to someone in \texttt{COUNTRY}, and belongs to the \texttt{CATEGORY} category. Please explain your reasoning step by step for identifying each entity. Please format your response as a JSON object where the keys are the culture-specific entities and the values are lists of potential replacements. The reasoning should be a separate key-value pair. The replacement list should be VERY EXHAUSTIVE and such that the replacements are highly relevant to \texttt{COUNTRY}. ONLY output the JSON, DO NOT WRITE ANYTHING ELSE."  \\

\textbf{GPT4 (Calculate Overlap)} \\
"You are given two lists of objects, List 1 and List 2. Your task is to identify objects that are similar between the two lists, allowing for minor variations in spelling, wording, or punctuation. Do not consider broad categories like 'animal', 'plant', 'flower', 'object', 'food', 'material', or any similar broad terms. Focus only on specific, meaningful matches between the lists, ignoring any broad or generic terms. Please explain your reasoning step by step and provide a final list of matching objects at the end. Provide the output as a JSON object with the following format. DO NOT WRITE ANYTHING ELSE." \\

\label{tab:prompts}
\end{tcolorbox}

\twocolumn

\onecolumn
\begin{tcolorbox}[colback=white,colframe=black,boxrule=0.5mm,arc=4mm,title=Prompts used to for VLM-scores]
\ttfamily 

\textbf{Gemini-1.5-Pro (Intent)} \\
"Given the input image and its category \texttt{CATEGORY}, determine if the comparison image belongs to the same category as the input. Please explain your reasoning step by step, and provide a final score at the end. The score should be between 1 and 5, where: 1 = Dissimilar Category, 5 = Same Category."\\
\emph{source-image} \\
\emph{target-image} \\
"The output should be a JSON object ONLY with the following format: \{"reasoning": ..., "score": number\}" \\

\textbf{Gemini-1.5-Pro (Creativity)} \\
"Given the input image and its category \texttt{CATEGORY}, determine if there are visual changes in the second image as compared to the first. Please explain your reasoning step by step, and provide a final score at the end. The score should be between 1 and 5, where: 1 = No visual change, 5 = High visual changes."\\
\emph{source-image} \\
\emph{target-image} \\
"The output should be a JSON object ONLY with the following format: \{"reasoning": ..., "score": number\}" \\

\textbf{Gemini-1.5-Pro (Cultural Relevance)} \\
"You will be given two images. You have to assess how culturally relevant both images are with respect to the culture of \texttt{COUNTRY}. Please explain your reasoning step by step for both images, specifically considering cultural symbols, styles, traditions, or any features that align with the culture of the speaking population of \texttt{COUNTRY}. For each image, the final score should be a number between 1 to 5, where 1 and 5 mean the following: 1 = Not culturally relevant, 5 = Culturally relevant."\\
\emph{source-image} \\
\emph{target-image} \\
"The output should be a JSON object ONLY with the following format:\{"first\_reasoning": ..., "first\_score": number, "second\_reasoning": ..., "second\_score": number\}" 
\label{tab:vlm_prompts}
\end{tcolorbox}
\twocolumn
\end{document}